\documentclass[conference]{IEEEtran}
\IEEEoverridecommandlockouts
\usepackage{cite}
\usepackage{amsmath,amssymb,amsfonts}
\usepackage{algorithmic}
\usepackage{graphicx}
\usepackage{textcomp}
\usepackage{xcolor}
\usepackage{mathtools}
\usepackage{float}

\usepackage[style=base]{caption}
\usepackage[utf8]{inputenc}
\usepackage[font=small,labelfont=bf]{caption} 

\definecolor{cb_red}{rgb}{0.89,0.1,0.11}
\definecolor{brown}{rgb}{0.59, 0.29, 0.0}
\DeclarePairedDelimiter\floor{\lfloor}{\rfloor}

\def\BibTeX{{\rm B\kern-.05em{\sc i\kern-.025em b}\kern-.08em
    T\kern-.1667em\lower.7ex\hbox{E}\kern-.125emX}}
\begin{document}
\pagenumbering{arabic}
\title{ Fully Automated Machine Learning Pipeline \\ for Echocardiogram Segmentation \\
}

\author{%
  \IEEEauthorblockN{%
    Thuy Hang Duong Thi\IEEEauthorrefmark{1} \IEEEauthorrefmark{2},
    Tuan Nguyen Minh\IEEEauthorrefmark{1} \IEEEauthorrefmark{2},
    Phi Nguyen Van\IEEEauthorrefmark{2} and
    Quoc Long Tran \IEEEauthorrefmark{2} \IEEEauthorrefmark{4}%
  }%
  \IEEEauthorblockA{\IEEEauthorrefmark{2} University of Engineering and Technology, Vietnam National University}%
}



%
    

\maketitle

\begingroup\renewcommand\thefootnote{\IEEEauthorrefmark{1}}
\footnotetext{Equal contribution}
\endgroup
\begingroup\renewcommand\thefootnote{\IEEEauthorrefmark{4}}
\footnotetext{Corresponding Author}
\endgroup

\begin{abstract}
Nowadays, cardiac diagnosis largely depends on left ventricular function assessment. With the help of the segmentation deep learning model, the assessment of the left ventricle becomes more accessible and accurate. However, deep learning technique still faces two main obstacles: the difficulty in acquiring sufficient training data and time consuming in developing quality models. In the ordinary data acquisition process, the dataset was selected randomly from a large pool of unlabeled images for labeling, leading to massive labor time to annotate those images. Besides that, hand-designed model development is strenuous and also costly. This paper introduces a pipeline that relies on Active Learning to ease the labeling work and utilizes Neural Architecture Search's idea to design the adequate deep learning model automatically. We called this Fully automated machine learning pipeline for echocardiogram segmentation. The experiment results show that our method obtained the same IOU accuracy with only two-fifths of the original training dataset, and the searched model got the same accuracy as the hand-designed model given the same training dataset.
\end{abstract}

\begin{IEEEkeywords}
  Active Learning, NAS, segmentation, Monte Carlo dropout, echocardiogram
\end{IEEEkeywords}

\section{Introduction}

Computer-aided medical diagnosis are one of the most active machine learning research domain in recent years.
Deep learning models have recently obtained higher performance in several image segmentation problems due to their representation power and generalization capability\cite{Unet}\cite{FCN}. Specifically, many medical image analysis applications rely on automated image segmentation.
\par

 However, to do supervised learning in deep learning, a model needs to be trained on data features with corresponding labels. After the model is trained and adequately tested, it can be used to make future predictions. There are two main problems in this case: the amount of labeled data and the model quality. Most of the time, labeled data is not available or only partially available. Because of that, a domain expert is required to do a manual job by labeling each of the instances separately. Another trade-off comes up since it is unclear whether annotating all the available data will benefit the model, considering the time and effort needed to do the work. The domain expert might label some recurring or very similar instances multiple times, which also creates more work and will not benefit the training process.\par
 
This phenomenon is often called the Curse of Big Data Labeling - much-unlabeled data is available while it often has abundant unlabeled data. Therefore, it can be costly, labor-intensive, and still ends up with problems like annotations missing from some genre or the labeled data distribution is biased or insufficient. \par
 
In this paper, we aim to solve those problems. Following is our main contributions for this work:
\begin{itemize}
    \item The design of a neural network architecture search for echocardiogram segmentation: 
    We provide an Auto Architecture Search and Auto Hyper-parameter Tuning method and a codebase which only needs a few lines of code to set up the search space with various searching sub-spaces. Numerous deep learning segmentation models were generated from the search space. The best found model obtained the same IOU accuracy as the manual developing model did. 
    \item Machine learning pipeline with Active Learning based (selecting data to annotate):
A machine learning model weight is generated automatically for echocardiogram segmentation given unlabeled pool data and a small initial labeled pool. Some images in the unlabeled pool data are selective without human intervention for the next training phase to make a better segmentation model. By now, human is assigned to label particular data, not to label random data as the usual way in supervised learning. We achieved a model with the same IOU accuracy with only two-fifths of data annotation as the well-tuned model trained with full data annotation.

\end{itemize}
    We called them \textit{Fully automated machine learning pipeline for echocardiogram segmentation.}

\section{Related work}

\subsubsection{Annotation burden decreasing}
Although deep learning's significant success in image segmentation tasks\cite{Unet}\cite{FCN}, deep learning-based segmentation algorithms still have a hard time getting enough training data due to the high cost of annotation. Weakly supervised segmentation algorithms\cite{transsemi} have been proposed to reduce the annotation burden in image segmentation tasks. However, the question of how to choose representative data samples for annotation is frequently disregarded. Active Learning, a form of semi-supervised machine learning, can be used as an annotation suggestion to query informative examples for annotation to solve this problem. In natural scene picture segmentation, good performance can be achieved with substantially less training data when utilizing Active Learning. In Biomedical Image Segmentation, with Active Learning-based, Haohan Li et all. \cite{haohan} achieved SOTA (at that moment) segmentation performance in MICCAI 2020 by using only a portion of the training data. However, their method used bootstrapping strategy which was time-costly and resource-consuming. Additionally, their proposed model needed to create new convolutional branches and many steps to estimate the image's characteristic to determine to select the image or not for labeling step.

\subsubsection{Automatic deep learning model developing }
 NAS (Neural architecture search\cite{nas}) is a method for creating artificial neural networks by automating the design process. NAS has been used to create networks that perform as well as or better than hand-built architectures. Nevertheless, if we use NAS directly in  our problem, there will be a bulk of search sub-spaces that we can not finish the searching process, as well as the hyper parameters have not yet been sought simultaneously.

\subsubsection{Echocardiography Image Segmentation}
Left ventricle segmentation is an essential step in Echocardiography, ultrasound imaging of the heart, to acquire qualitative measurements such as the left ventricle's location and for quantitative measurements such as length, area, volume, or the ejection fraction index, an important indicator of how well the heart is. With the recent advances in supervised deep learning in echocardiography\cite{segmentation1, segmentation2, segmentation3, segmentation4, segmentation5, segmentation6, segmentation7}, it is now possible to achieve the expert accuracy with the right quantity and quality of data.

\section{Proposed Method}

To ease the labeling working and create effective deep learning models, we propose a fully automated machine learning pipeline (from data collection-annotation (III.C) to develop the adequate model (III.B) automatically).

\subsection{Baseline}

The most famous Convolutional Networks for Biomedical Image Segmentation is Unet\cite{Unet}. The network consists of a contracting path and an expansive path, which gives it the U-shaped architecture. The contracting path is a typical convolutional network that consists of repeated application of convolutions, each followed by a rectified linear unit (ReLU) and a max pooling operation. Using this Unet based, after developing many models, we obtained the best IOU accuracy value which equal to 87\%. However, those hand-designed networks are time expensive and take a lot of work to develop. Hence, the need to develop deep learning model in an automatic way is required. \par

The training dataset was chosen at random from a massive pool of unlabeled images that needed to be labeled. There was lots of data and a lot of labor involved in labeling these images.
Selecting data randomly for annotating might end up with problems like annotations missing from some genre, or the labeled data distribution is bias or not sufficient.\par

\subsection{Finding the best segmentation model with Encoder Decoder Neural Architecture Search (EDNAS)}
\textbf{Models architecture}\par
All segmentation models in  our segmentation model are made of:\par
\textbf{encoder} (feature extractor, also known as backbone): Encoder is a “classification model” which extract features from image and pass it to decoder. There is an encoder-depth which is a number of stages used in encoder in range [3, 5]. Each stage generates features two times smaller in spatial dimensions than previous one (e.g. for depth 0 we will have features with shapes [(N, C, H, W)], for depth 1 - [(N, C, H, W), (N, C, $\floor{H / 2}$, $\floor{W / 2}$)] and so on;  (N, C, H, W)  means a feature whose layout is (batch size, channel, height, width)). \par
\textbf{decoder} (features fusion block to create segmentation mask): From a list of extract features in encoder's output, decoder builds the output mask. \par
\textbf{segmentation head} (final head to reduce the number of channels from decoder and upsample mask to preserve input-output spatial resolution identity)\par
In  our segmentation image work, \textit{EDNAS is characterized} as a system with \textit{three major components}:\par

\begin{figure}
    \centering
    \captionsetup{justification=centering}
    \includegraphics[width=0.47\textwidth]{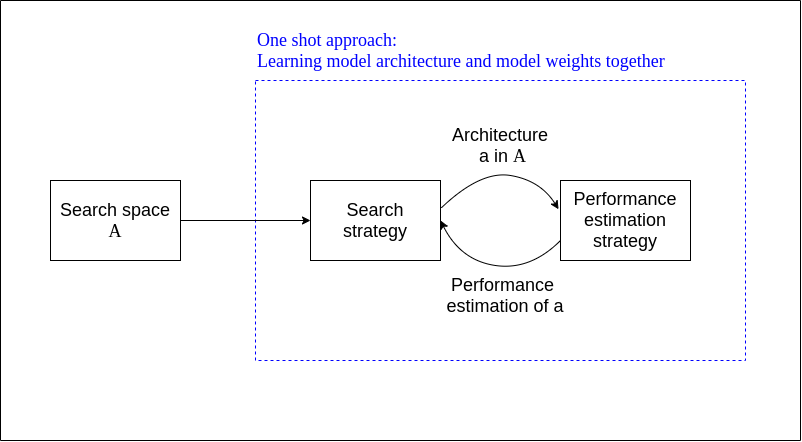}
    \caption{Three main components of Encoder Decoder Neural Architecture Search (EDNAS) models. }
    \label{fig:nas}
\end{figure}

\begin{itemize}
    \item Search space: The EDNAS search space defines a set of encoder names (which are also backbone and popular classification models) (e.g. ResNet, ResNeXt, DenseNet), a set of architecture models (e.g: Unet, Feature Pyramid Network FPN, DeepLab family).
    
    \item Search algorithm: A EDNAS search algorithm samples a population of network architecture candidates. It receives the child model performance metrics as rewards (e.g. high accuracy, low latency) and optimizes to generate high-performance architecture candidates. \par
    
    
        Random search is the most naive baseline and was used in this work. It samples a valid architecture candidate from the search space at random and no learning model is involved. Random search has proved to be quite useful in hyper-parameter search. Because our search space is well-designed so that random search could be a very challenging baseline to beat.

    \item Evaluation strategy: We need to measure, estimate, or predict the performance of a large number of proposed child models in order to obtain feedback for the search algorithm to learn. The process of candidate evaluation could be very expensive and many new methods have been proposed to save time or computation resources. In  our work, performance estimation is based on computing IOU metric in testing dataset

\end{itemize}

\begin{figure*}[ht]
    \centering
    \includegraphics[width=0.75\textwidth]{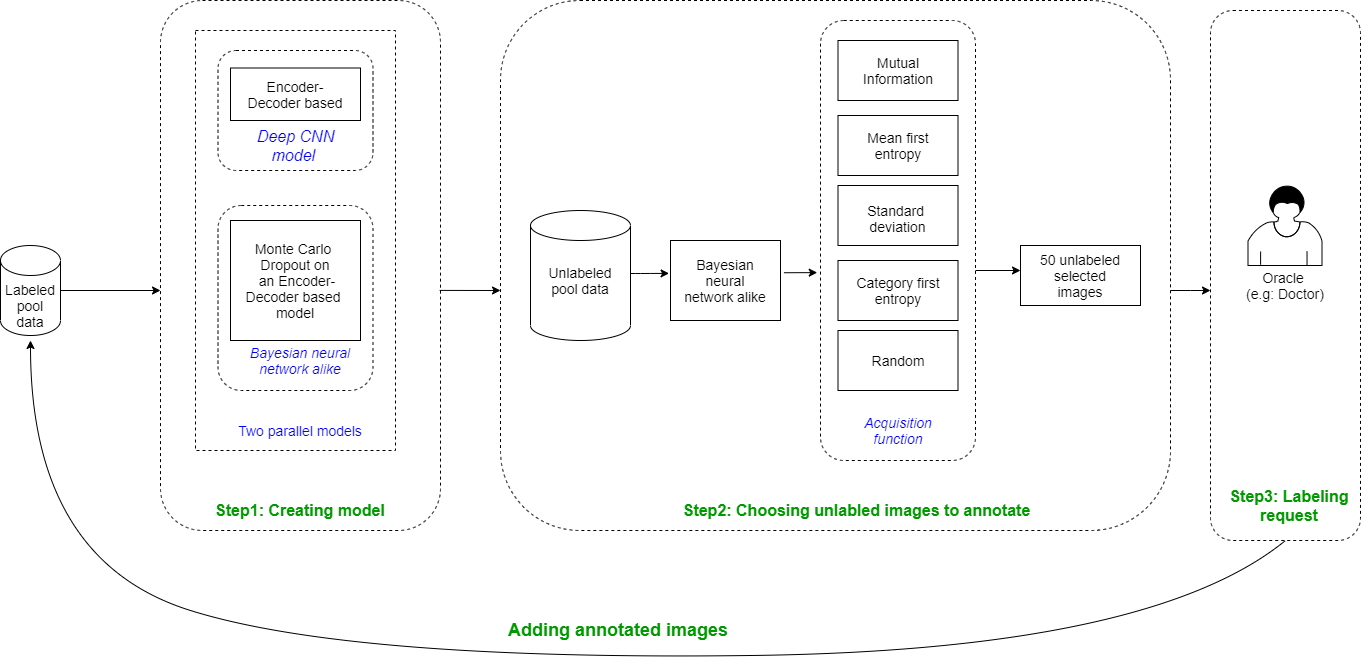}
    \captionsetup{justification=centering}
    \caption{Echocardiogram segmentation pipeline based on Active Learning}
    \label{fig:pipeline}
\end{figure*}



\subsection{Active Learning: Same accuracy with two-fifths data annotation} 
Figure \ref{fig:pipeline} describes our  workflow of echocardiogram segmentation pipeline based on Active Learning with the Pool-based sampling scenario which assumes that all of the original labeled data is unlabeled. The reason is, in  our experiment, there are a lot of queries for annotating, and the doctors are not free all the time, also the cost attached to it is enormous.\par 
In each phase, firstly, two models are training parallel (one for suggesting which unlabeled images should be annotated and one for obtaining accurate segmentation). Secondly, we pass every single image in unlabeled pool data into our Bayesian neural network alike which then inferring T probability masks, and using acquisition function we can get image's uncertainty values. Finally, images that have the top highest uncertainty values will be selected to label. The process continues with the next phase, this makes a loop of  model training and data choosing. The stopping criteria: when the performance reaches a threshold (87\% on IOU metric) or run out of budget to do the annotation (in our experiment the budget is the amount of labeled data).\par 

\subsubsection{Uncertainty estimation}
For the Neural network combined with Convolutional layers, the architecture is deep, it is computationally impossible to estimate the model uncertainty for marginalizing every possible weight in the layers. So we use approximate inference in Bayesian Neural Network. The advantage of Monte Carlo Dropout is that the model will not be changed when implementing the algorithm, and only the results after several inferences will be collected. The Monte Carlo Dropout is a way to sample for deep models.\par

 Because of the similarity of the Bernoulli distribution and dropout technique in shutting down the node at the setting probability, the algorithm is regularizing the network during training with dropout and keeping the dropout layers at inference to draw Monte Carlo samples with different dropout masks as approximate.\par
For the sake of time and resources, we implemented the CNN network with only adding dropout after the last convolution layer and all decoder convolution layers. Furthermore, the network is quite complex. As suggested by Yarin Gal in his PhD thesis\cite{yarin_thesis}, adding dropout layer after all convolutional layers or adding dropout after the last convolutional layer has the same purpose, as he presented both those two experiments. In this work, we  inferred T(=30 in  our experiment) stochastic forward passes through the model following the Bayesian interpretation of dropout. \par

\begin{figure}[H]
    \centering
    \captionsetup{justification=centering}
    \includegraphics[width=0.45\textwidth]{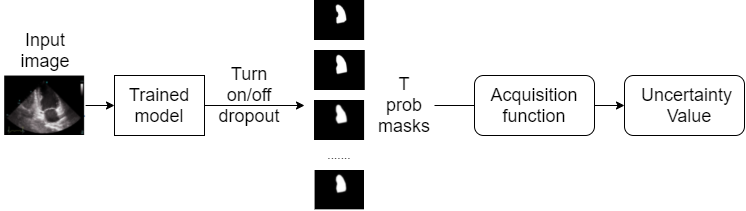}
    \caption{The overview of estimating image's uncertainty}
    \label{fig:est_uncer}
\end{figure}

\subsubsection{Acquisition function}
Function mapping probability masks to uncertainty value.





From uncertainty estimation, we can obtain the predictive probability to calculate the value of information. For each image in the unlabeled pool, accumulating the uncertainty computed for all the pixels can give the value measuring uncertainty. The samples queried will be used to expand the training set. The query function are mainly based on calculating the entropy. ``T" in below formula means the number of inferences from Monte Carlo Dropout of our deep learning model. \par

Given x as the input image, W$_t$ is the trained weight with the corresponding inference t$^{th}$, y is the output mask, p is the probability function, C is the number of possible classes (C = 2 in our problem, 0 and 1, 0: when there is no left ventricular, 1: otherwise). As written in Jiarui Xiong's thesis \cite{stockholm}, there were five acquisition functions:

\begin{itemize}
    \item Category First Entropy\par
This query function calculates the entropy between all the classes of one pixel first in every single mask, then averages that value in multiple T probability masks. The entropy uncertainty acquisition function is:
\begin{equation}
\begin{array}{c}
\mathcal{H}_{category}
=-\frac{1}{T} \sum_{t=1}^{T} \sum_{c=1}^{C} [p\left(y=c \mid \mathbf{x}, \mathbf{W}_{t}\right) * \\ \log p\left(y=c \mid \mathbf{x}, \mathbf{W}_{t}\right)]
\end{array}
\end{equation}

If $\mathcal{H}_{category}$ is small (close to zero), it means the entropy of each model is small, then every individual model is confident about its prediction, however, not all models give the same prediction. On the other side, maybe each model gives us different class predictions with high confidence.
    \item Mean First Entropy\par
This query function is extracting the mean of T pixels from T probability masks first, hence, we have the classes probability of one pixel, then calculating the entropy for that pixel, we got the final result.
\begin{equation}
\begin{array}{c}
p(\mathbf{y} \mid \mathbf{x}) \approx \frac{1}{T} \sum_{t=1}^{T} p\left(\mathbf{y} \mid \mathbf{x}, \mathbf{W}_{t}\right)\\=\frac{1}{T} \sum_{t=1}^{T} \operatorname{softmax}_{\left(\mathbf{W}_{t}\right)}(\mathbf{x}) \\
\mathcal{H}_{mean}=-\sum_{c=1}^{C} p(y=c \mid \mathbf{x}) \log p(y=c \mid \mathbf{x})
\end{array}
\end{equation}
If $\mathcal{H}_{mean}$ is small (close to zero), we can infer that every model predicts the same class with high assurance.  
    \item Mutual Information\par
This query function calculates the difference of two calculated entropy above.
$$
\mathcal{H}_{mi} = |\mathcal{H}_{mean}-\mathcal{H}_{category}|
$$
It can be regarded that if $\mathcal{H}_{mi}$ is small (close to zero), it means all the inferred stochastic models give the same classifying result.

    \item Standard deviation\par
This query function uses the standard deviation of the predictive probability from multiple models.
$$
\operatorname{STD}\left(p_{T}\right)
$$
If $\operatorname{STD}\left(p_{T}\right)$ is nearly zero, all the models produce the same class probability, however, it can be high uncertainty in each model's result.

    \item Random \par
Using this acquisition function is equivalent to choosing a point uniformly at random from the pool.

\end{itemize}

Now, with the obtained uncertainty values, we can get the top K (e.g., K = 50) images with the highest values (the higher value the more uncertain image). These chosen data will be labeled by the Oracle (the doctors). Latter, they are added to the labeled pool data.

\section{Experiments}

\subsection{Dataset}
Our own dataset contains 3427 pairs (image, mask) from 252 videos drawn from The 2 Chamber View in Echocardiography. The data was annotated by experts. The resulting images were in variety range of resolution: 768x1024, 434x636, 422x636 (pixels).\par

\begin{itemize}
    \item For finding the best segmentation model with EDNAS:
    We used 2690 pairs (from 118 videos) in training set and 737 pairs (from 64 videos) in testing set.
    \item For active Learning: The initial training set contains 40 pairs (image, mask) from multiple videos. There are 737 pairs (from 64 videos) in test set in every Active Learning phase.
\end{itemize}
 For the data pre-processing, every image is resized into a fixed resolution: 256x256 pixels, pixel value is scaled into range [0,1] in the preprocessing step, then applying a sequence of augmentation transformations which are Random rotation, flip, transpose, GaussNoise with probability 0.2, motion blur, median blur, blur, optical distortion, grid distortion, IAAPiecewiseAffine, CLAHE, random brightness contrast. All the masks, which are prepared by the doctors with the help of a labeling system, are images containing 0 and 1 pixel value. \par

\subsection{Implementation detail}

\subsubsection{Active Learning: Same accuracy with two-fifths data annotation}
Firstly, we had 40 labeled-images for initial training data pool (As Active Learning data assumption we mention in Section III.C). Using method in our section III.B, we found out 2 model architectures, namely, Pan-RegnetY120, DeepLabV3-DPN98, had the fastest training time and highest IOU accuracy accordingly.\par
For each phase, newly labeled images will be added to training set. Note that uncertainty value is calculated for every single image in the unlabeled pool and all elements in the list of those uncertainty values will be sorted in descending order. Fifty images corresponding with fifty maximum values will be given to the doctor to label, latter being added to training pool data. Two models, one model that is embedded drop out is used for evaluating uncertainty, the other which does not use drop out is used for segmentation, are trained parallel every phase.\par

Therefore, there are \(40 + 50(i-1) = 50i -10\) images in training pool at $i$th phase. Within a phase, we trained the model in 30 epochs.\par

Stopping criterion: Active Learning process will be stopped when the IOU score in test set obtain approximately the threshold 87\% (which is the best one received with the full dataset). \par

\subsubsection{Finding the best segmentation Model with EDNAS}
As we presented in the previous chapter, the model which does the segmentation job is based on decoder-encoder network. 
The encoder is drawn from these networks: Resnet, ResNeXt, ResNet(X/Y), SE-Net, SK-ResNe(X)t, Dense Net, Efficient Net, DPN, and their extended versions with a variety configurations \cite{search}. The network architecture is drawn from: Unet, Unet++, Linknet, FPN, PSPNet, PAN, DeepLabV3, DeepLabV3+ \cite{search}.\par
The learning rate set: $\{1 e-4 ; 4 e-4 ; 1 e-5 ; 5 e-5 ; 1 e-6 ; 4 e-6\}$. \par 
Batch size set: $\{4 ; 8 ; 16\}$. \par
In total, the search space contains:
$$ 46*9*6*3 = 7452 \text{ subspaces} $$
 Where:  46 - the number of encoder architectures; 9 - the number of decoder architectures; 6 - the number of learning rates; 3 - the number of batch sizes. \par
 In our experiment, Nvidia GEFORCE RTX 2080 was used, training time for searching the best fitted model is about 6 days. 
 

\subsection{Result}
\subsubsection{Finding the best segmentation model with EDNAS}

\begin{table*}[ht]
\centering
\caption{Performance rank between the best 5 searched models}
\label{tab:best5models}
\begin{tabular}{|l|l|l|l|l|l|}
\hline
Rank & IOU metric & batch size & learning rate & Encoder                & Architecture \\ \hline
1st  & 87.00 \%   & 4          & 4e-5          & timm-skresnet34        & Linknet      \\ \hline
1st  & 86.95 \%   & 4          & 1e-5          & timm-regnety-120       & PAN          \\ \hline
3rd  & 86.85 \%   & 8          & 5e-5          & timm-resnest50d-1s4x24d & UnetPlusPlus \\ \hline
4th  & 86.75 \%   & 8          & 5e-4          & vgg16-bn               & UnetPlusPlus  \\ \hline
5th  & 86.68 \%   & 8          & 1e-5          & se-resnext50-32x4d     & DeepLabV3    \\ \hline
BASELINE &   87\%  &           &               &                        &  3D ResUNet    \\ \hline
\end{tabular}
\end{table*}

Table \ref{tab:best5models} is the result from a bulk of training multiple models trials with various hyperparameter configurations. We received two models which get the first rank and their performances are 87\% on the intersection over union metric with approximately 2690 images of the training dataset.

The IOU accuracy range is 86.68\% to 87.0\%. As can be seen from the result, the encoder architecture almost belongs to the Resnet family. In conclusion, we can see that the incredible power of searching and automation work. \par


\subsubsection{Active Learning on PAN-RegnetY120}
We use two dropout adding strategies: segmentation head only (which means dropout layer is added after the last convolutional layer), and full-decoder layer (which means dropout layers are added after every convolutional layer in decoder-part). 

\begin{figure}[htbp]
    \centering
    \includegraphics[width=0.5\textwidth]{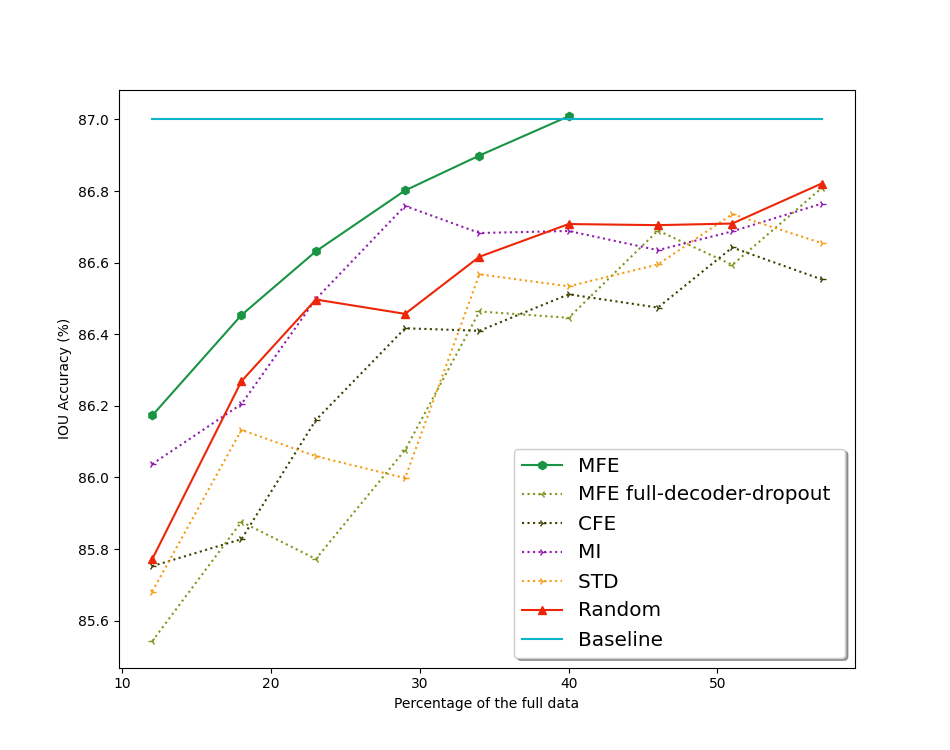}
    \caption{PAN RegNetY120 model IOU accuracy with different data acquisition, where baseline value is the best IOU accuracy obtained by the base model}
    \label{fig:all_no}
\end{figure}

The models in Fig. \ref{fig:all_no} (which are named in the form of AcquisitionFunctionName, having  PAN-RegNetY120 architecture) are the models trained by turning off dropout within each Active Learning phase. The result of MFE, CFE, MI, STD, Random was the average of three experiment's outcomes with different initialization seed. \par
Comparison among all the query strategy functions is shown in Fig.\ref{fig:all_no}. MFE has superior performance compared to Random. While MI has better performance than Random during the first 16 Active Learning phases, the remainder phases perform slightly worse. STD's result and CFE's result are not good as Random's one. \par

For details, in table \ref{tab:5phases_in_act}, it is surprising that in \textit{less than 5 phases} (using 240 images, accounting for 9\% of the original), almost models \textit{obtain 85\% IOU accuracy} on the test set. Even with \textit{only 90 images} on training set, STD model gets \textit{82.46\%} IOU accuracy or \textit{85\%} IOU accuracy with training on \textit{190 images}. This can be explained by the similarity among data or the evaluation metric is not exact enough for this task. \par
As the same as STD does, Random and MI gets similar result in the first five phases. It can be drawn that using \textit{Mean first entropy} (MFE) acquisition function, we gain the best performance which is \textit{ the expected accuracy} (above eighty seven percentage) in the \textit{21st phase} in which the data accounts for nearly 40\% original training dataset.  \par

\begin{table*}[ht]
\centering
\captionsetup{justification=centering}
\caption{PAN RegNetY120 model IOU accuracy and percentage result.  The first five phases are listed in details.  Other phase is the phase which the model gets its best performance.}
\label{tab:5phases_in_act}
\begin{tabular}{|c|c|c|c|c|c|c|}
\hline
Phase & The percentage of original dataset & \multicolumn{5}{c|}{\begin{tabular}[c]{@{}c@{}}Test IOU accuracy with\\  corresponding acquisition funciton\end{tabular}} \\ \cline{3-7} 
\multicolumn{1}{|l|}{}                       &                                                     & STD                & Random             & MI                 & MFE           & CFE               \\ \hline
1                                            & 1.6\% (1/64, 40 imgs)                               & 67.6\%             & 68.32\%            & 68.37\%            & 64.20\%                & 67.16\%           \\ \hline
2                                            & 3.3\% (90 imgs)                                     & 82.46\%            & 83.7\%             & 83.3\%             & 78.64\%                & 81.97\%           \\ \hline
3                                            & 5.2\% (140 imgs)                                    & 84.19\%            & 84.7\%             & 84.23\%            & 81.83\%                & 83.91\%           \\ \hline
4                                            & 7.1\% (190 imgs)                                    & 85.06\%            & 85.4\%             & 84.73\%            & 83.0\%                 & 84.94\%           \\ \hline
5                                            & 9\% (240 imgs)                                       & 85.71\%            & 85.56\%            & 85.4\%            & 83.88\%                & 84.87\%           \\ \hline
18                                           & 33.1\% (890 imgs)                                   & 86.24\%             &  86.54\%           &   \textbf{86.78}\%          & 86.86\%                & \textbf{86.84\%}           \\ \hline
21                                           & 38.7\% (1040 imgs)                                   & 86.48\%            &  86.70\%           &  86.58\%           & \textbf{87.02\%}                & 86.36\%       \\ \hline
29                                           & 53\% (1440 imgs)                                     & 86.62\%            &  86.66\%             &  85.44\%          &    \textbf{87.08\%}               & 86.50\%               \\ \hline
31                                           & 57\% (1540 imgs)                                     & \textbf{86.65\%}             & \textbf{86.83\%}            & 86.76\%            &  87.01\%              &  86.55\%           \\ \hline
\end{tabular}
\end{table*}

\subsubsection{Active Learning on DeepLabV3-DPN98}
\begin{figure}[htbp]
    \centering
    \includegraphics[width=0.5\textwidth]{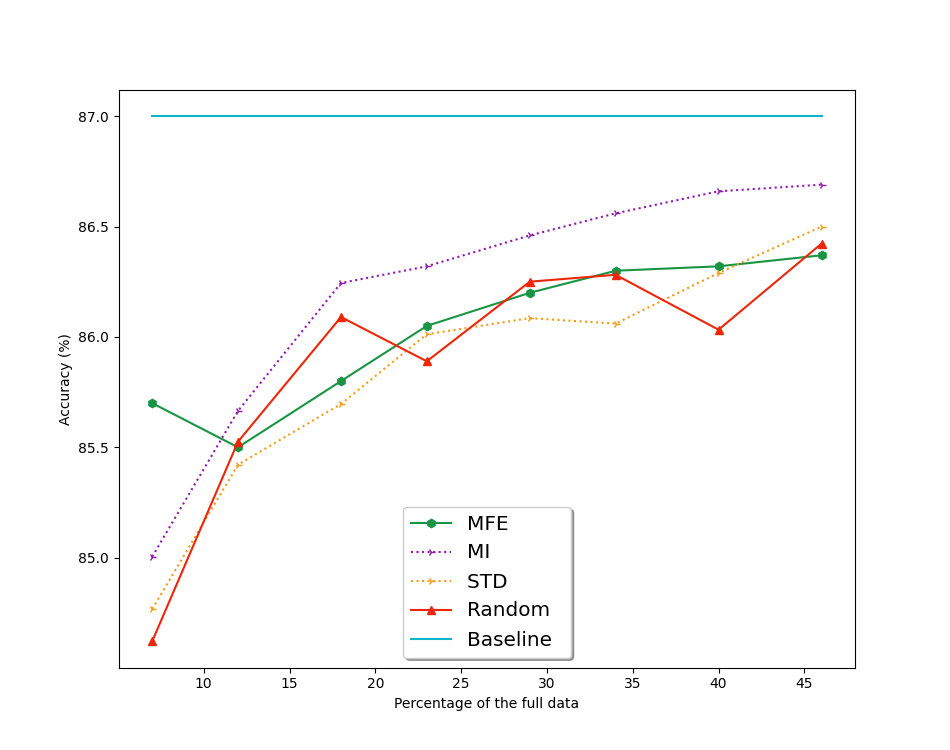}
    \caption{Model IOU accuracy with different data acquisition on DeepLabV3-DPN98, where baseline value is the best IOU accuracy obtained by the base model}
    \label{fig:dp_98}
\end{figure}

With DeepLabV3-DPN98 as the model architecture, and dropout was added after the segmentation head (the last convolution layer). Fig \ref{fig:dp_98} is the result of Active Learning through phases.\par

In this experiment, we tried 26 Active Learning phases with 4 different acquisition functions, namely: MFE, MI, STD, Random (due to the limit of time and resource we did not try all 5 mentioned acquisition functions). Unlike the PAN RegNetY120's result, through 26 phases, in terms of performance, MI is always better than Random while MFE is similar to Random. This can be explained by huge difference between PAN-RegNetY120 params (50,290,512) and DeepLabV3-DPN98 params (79,758,785) ( DeepLabV3-DPN98 params is one point six times more than PAN-RegNetY120 params), which created different Bayesian neural network styles when using Dropout. Compare to PAN-RegnetY120 accuracy (87\%), DeepLabV3-DPN98 accuracy (86.6\%) is slightly lower, both using two-fifths of original training data.  \par
In both two architectures, MI's performance and MFE's performance are equal or greater than Random's one.
In reality, we prefer to try on all of the acquisition functions in some first phases in Active Learning and then choosing the best acquisition function to do the rest of the task.
\section{CONCLUSION}
We offer a fully automated machine learning pipeline consisting of: (1) an encoder-decoder base model utilizing NAS's idea that finds the best architecture and configuration for echocardiogram segmentation, and this model had the same accuracy as the prior manual deep learning had; (2) an Active Learning based machine learning pipeline, which progressively reduces the burden of manual labeling in biomedical image segmentation. With the use of only two-fifths of the annotated training data, our model achieved the same accuracy. In the future, we will try on more dataset and metric, reduce the similarity between the selective image in order to expand the diversity of collected data, find better ways to estimate image uncertainty, try to apply in other fields in deep learning such as object tracking, natural language processing.
\section*{Acknowledgment}
This work has been supported by VNU University of Engineering and Technology.

\end{document}